\documentclass[a4paper,oneside,12pt,english]{article}
\usepackage{url}
\usepackage[pdftex]{color,graphicx}
\usepackage[backref=page]{hyperref}
\hypersetup{plainpages = false,
              breaklinks=true,
              a4paper=true,
              linktocpage,
              pdfauthor={Niko Brummer},
              pdfstartpage=1, 
              pdfstartview=FitH,  
              pdfkeywords={}}

\usepackage{amsmath}%
\usepackage{amsfonts}%
\usepackage{amssymb}%
\usepackage{graphicx}

\DeclareMathOperator*{\argmax}{argmax}

\begin{document}
\title{The EM algorithm and the Laplace Approximation}
\author{Niko Br\"ummer\\AGNITIO Research South Africa}
\maketitle
The Laplace approximation calls for the computation of second derivatives at the likelihood maximum. When the maximum is found by the EM algorithm, there is a convenient way to compute these derivatives. The likelihood gradient can be obtained from the EM-auxiliary, while the Hessian can be obtained from this gradient with the Pearlmutter trick. 

\section{The Laplace approximation}
\def\hess{\boldsymbol{\Lambda}}
\def\ND{\mathcal{N}}
Let $X$ denote the observed data, $H$ some hidden variables and $\Theta$ the model parameters. We assume the \emph{joint distribution}: 
\begin{align}
P(X,H,\Theta) = P(X|H,\Theta) P(H|\Theta) P(\Theta) 
\end{align}
is easy to work with, while the \emph{marginal distribution}:\footnote{All integrals are definite integrals, with fixed boundaries. If $H$ is discrete, the integral can be replaced by summation.}
\begin{align}
P(X,\Theta) &= \int P(X,H,\Theta) \,dH
\end{align}
has a more complex form. The Laplace approximation\footnote{See: Christopher M. Bishop, \emph{Pattern Recognition and Machine
Learning} (Information Science and Statistics), Springer, 2007; David J. C. MacKay, \emph{Information Theory, Inference, and Learning Algorithms}, Cambridge University Press, 2003.} calls for finding the \emph{mode} and \emph{Hessian}, w.r.t.\ $\Theta$:
\begin{align}
\hat\Theta &= \argmax P(X,\Theta), &\text{and}&&
\hess &= \nabla^2\log P(X,\hat\Theta) 
\end{align}
The approximation is:
\begin{align}
P(\Theta|X) \approx \ND(\Theta|\hat\Theta,-\hess^{-1})
\end{align}  

\section{EM-algorithm}
If we are using the EM-algorithm for finding the maximum, $\hat\Theta$, then the EM-auxiliary provides a convenient route to the Hessian.

\subsection{The EM auxiliary}
Let $\Theta'$ be any valid parameter value satisfying $\int P(H|X,\Theta')\,dH=1$. We construct the EM auxiliary as follows:
\begin{align}
\begin{split}
\log P(X,\Theta) &= \int P(H|X,\Theta') \log P(X,\Theta) \,dH \\
&= \int P(H|X,\Theta') \log P(X|\Theta) \,dH + \log P(\Theta)\\
&= \int P(H|X,\Theta') \log \frac{P(X,H|\Theta)}{P(H|X,\Theta)} \,dH +\log P(\Theta)\\
&= \int P(H|X,\Theta') \log \left[
\frac{P(X,H|\Theta)}{P(H|X,\Theta)} 
\times\frac{P(H|X,\Theta')}{P(H|X,\Theta')} 
\right]\,dH +\log P(\Theta)\\
&= A(\Theta',\Theta) + D(\Theta',\Theta)
\end{split}
\end{align}
where $A(\Theta',\Theta)$ is the \emph{EM-auxiliary}:
\begin{align}
A(\Theta',\Theta) &= \int P(H|X,\Theta') \log \frac{P(X,H|\Theta)}{P(H|X,\Theta')} \,dH 
+\log P(\Theta)
\end{align}
and  $D(\Theta',\Theta)\ge0$ is KL-divergence:
\begin{align}
D(\Theta',\Theta) &= \int P(H|X,\Theta') \log \frac{P(H|X,\Theta')}{P(H|X,\Theta)} \,dH 
\end{align}
Notice that if we zero the divergence by choosing $\Theta'=\Theta$, then:
\begin{align}
\log P(X,\Theta) &= A(\theta,\theta)
\end{align}

\subsection{Algorithm}
Although this note is not about the algorithm itself, we very briefly summarize it. An iteration of the EM-algorithm proceeds as follows: Start at $\Theta_1$. The E-step effectively maximizes $A(\Theta',\Theta_1)$ w.r.t.\ $\Theta'$ by simply setting $\Theta'=\Theta_1$, which minimizes (and therefore zeros) the divergence.\footnote{Here we vary $\Theta'$, while $\Theta$ and therefore $\log P(X,\Theta)$ remain \emph{fixed}. Then decreasing $D$ must increase $A$.} The M-step now maximizes $A(\Theta_1,\Theta)$, w.r.t.\ the \emph{other} parameter, $\Theta$, usually by zeroing partial derivatives. This gives some value $\Theta_2$, such that $A(\Theta_1,\Theta_2)\ge A(\Theta_1,\Theta_1)$. The net effect of both steps is:
\begin{align}
\log P(X,\Theta_2) = A(\Theta_2,\Theta_2) \ge A(\Theta_1,\Theta_2) \ge A(\Theta_1,\Theta_1) = \log P(X,\Theta_1)  
\end{align}

\section{Derivatives}
\def\ddt{\frac{\partial}{\partial\theta}}
We find the Hessian of $\log P(X,\Theta)$ in two steps. First we find the gradient, which we then  differentiate again using the Pearlmutter trick.

\subsection{Gradient}
The gradient of $\log P(X,\Theta)$ coincides with the gradient of the auxiliary. We show how this works.

Let $\theta$ denote some component of $\Theta$, then, for any value of $\Theta'$, we have:
\begin{align}
\ddt\log P(X,\Theta) &= \ddt A(\Theta',\Theta) + \ddt D(\Theta',\Theta)
\end{align} 
Note: we are differentiating \emph{only} w.r.t. the components of $\Theta$ and \emph{not} w.r.t.\ those of $\Theta'$. The derivative of the divergence is:
\begin{align}
\ddt D(\Theta',\Theta) &= -\int \frac{P(H|X,\Theta')}{P(H|X,\Theta)} \ddt P(H|X,\Theta)\,dH 
\end{align}
which conveniently vanishes at $\Theta'=\Theta$:
\begin{align}
\begin{split}
\left[\ddt D(\Theta',\Theta)\right]_{\Theta'=\Theta} &= -\int \ddt P(H|X,\Theta)\,dH \\
&= -\ddt \int P(H|X,\Theta)\,dH =  -\ddt 1 = 0
\end{split}
\end{align}
Putting this together, we find:
\begin{align}
\label{eq:grad}
\begin{split}
\ddt\log P(X,\Theta) &= \left[\ddt A(\Theta',\Theta)\right]_{\theta'=\theta} \\
&= \int P(H|X,\Theta)\ddt\log P(X,H,\Theta) \,dH
\end{split}
\end{align}
For exponential family distributions, the RHS is usually more convenient than the LHS, because now the log directly simplifies $P(X,H,\Theta)$. Also note that it is unnecessary to differentiate the  posterior $P(H|X,\Theta)$, or any associated entropy or divergence.

\subsubsection{Other derivatives}
\def\ddtp{\frac{\partial}{\partial\theta'}}
Just for interest, we mention here that there are two other derivatives that also vanish:
\begin{align}
\left[\ddtp A(\Theta',\Theta)\right]_{\Theta'=\Theta}
&= \left[\ddtp D(\Theta',\Theta)\right]_{\Theta'=\Theta}
= \left[\ddt D(\Theta',\Theta)\right]_{\Theta'=\Theta} = 0
\end{align}
where $\theta'$ is any component of $\Theta'$. This is because at $\Theta'=\Theta$, $A$ is maximized w.r.t.\ $\Theta'$, while $D$ is minimized w.r.t.\ both arguments. Only $\ddt A$ does not vanish here, because it is not necessarily at the maximum w.r.t.\ $\Theta$.

\subsection{Hessian}
\def\ddij{\frac{\partial^2}{\partial\theta_i\partial\theta_j}}
\def\ddi{\frac{\partial}{\partial\theta_i}}
\def\ddj{\frac{\partial}{\partial\theta_j}}
We first examine the Hessian analytically. We now consider $\theta_i,\theta_j$, both components of $\Theta$ and differentiate first w.r.t.\ the one and then the other:
\begin{align}
\begin{split}
&\ddij\log P(X,\Theta) \\
&= \ddj \int P(H|X,\Theta) \ddi\log P(X,H,\Theta)\,dH \\
&= \int P(H|X,\Theta) \ddij\log P(X,H,\Theta)\,dH \\
&\;\;\;\;+ \int \ddj P(H|X,\Theta) \ddi\log P(X,H,\Theta)\,dH \\
&= \left[\ddij A(\Theta',\Theta)\right]_{\Theta'=\Theta}  + \int \ddj P(H|X,\Theta) \ddi\log P(X,H,\Theta)\,dH \\
\end{split}
\end{align}
This is the Hessian of the auxiliary plus an extra term that can get messy to derive and implement. The Pearlmutter trick gives a convenient alternative:

\subsubsection{Pearlmutter trick}
\def\vvec{\mathbf{v}}
Let $\nabla f(\Theta)$, a column vector, denote the gradient of some multivariate function $f$, evaluated at $\Theta$. Similarly, let $\nabla^2f(\Theta)$, a square matrix, denote the Hessian. Then the Pearlmutter trick\footnote{Barak A. Pearlmutter, ``Fast exact multiplication by the Hessian'',
Neural Computation, vol. 6, pp. 147–160, 1994.} computes the product of the Hessian with an arbitrary column vector, $\vvec$, as: 
\begin{align}
\nabla^2f(\hat\Theta)\vvec &= \left[
\frac{\partial}{\partial\alpha}\nabla f(\hat\Theta+\alpha\vvec)
\right]_{\alpha=0}
\end{align}
When $\Theta$ has $n$ components, the trick must be applied $n$ times, to map out the columns of the Hessian by successively choosing $\vvec=[1,0,0,\ldots]$, $\vvec=[0,1,0,\ldots]$ and so on.

For practical implementation, the gradient using~\eqref{eq:grad} could be derived\footnote{The M-step should be based on those same derivatives.} and coded by hand. When that function is available, the differentiation could be done via forward-mode, algorithmic differentiation. If complex arithmetic is available, then that can be done with minimal coding effort via complex-step differentiation.

\end{document}